\documentclass[letterpaper]{article} 
\usepackage[submission]{aaai23}  
\usepackage{times}  
\usepackage{helvet}  
\usepackage{courier}  
\usepackage[hyphens]{url}  
\usepackage{graphicx} 
\usepackage{natbib}  
\usepackage{caption} 
\usepackage{algorithm}
\usepackage{algorithmic}
\usepackage{amsmath}
\usepackage{multirow}

\usepackage{newfloat}
\usepackage{listings}
\DeclareCaptionStyle{ruled}{labelfont=normalfont,labelsep=colon,strut=off} 
\floatstyle{ruled}
\newfloat{listing}{tb}{lst}{}
\floatname{listing}{Listing}
\title{ Semantic-Human: Neural Rendering of Humans from Monocular Video with Human Parsing}
\author{
    Written by AAAI Press Staff\textsuperscript{\rm 1}\thanks{With help from the AAAI Publications Committee.}\\
    AAAI Style Contributions by Pater Patel Schneider,
    Sunil Issar,\\
    J. Scott Penberthy,
    George Ferguson,
    Hans Guesgen,
    Francisco Cruz\equalcontrib,
    Marc Pujol-Gonzalez\equalcontrib
}
\affiliations{
    \textsuperscript{\rm 1}Association for the Advancement of Artificial Intelligence\\
    1900 Embarcadero Road, Suite 101\\
    Palo Alto, California 94303-3310 USA\\
    publications23@aaai.org
}

\usepackage{bibentry}

\begin{document}

\maketitle

\begin{abstract}
The neural rendering of humans is a topic of great research significance. However, previous works mostly focus on achieving photorealistic details, neglecting the exploration of human parsing. Additionally, classical semantic work are all limited in their ability to efficiently represent fine results in complex motions. Human parsing is inherently related to radiance reconstruction, as similar appearance and geometry often correspond to similar semantic part. Furthermore, previous works often design a motion field that maps from the observation space to the canonical space, while it tends to exhibit either underfitting or overfitting, resulting in limited generalization. In this paper, we present Semantic-Human, a novel method that achieves both photorealistic details and viewpoint-consistent human parsing for the neural rendering of humans. Specifically, we extend neural radiance fields (NeRF) to jointly encode semantics, appearance and geometry to achieve accurate 2D semantic labels using noisy pseudo-label supervision. Leveraging the inherent consistency and smoothness properties of NeRF, Semantic-Human achieves consistent human parsing in both continuous and novel views. We also introduce constraints derived from the SMPL surface for the motion field and regularization for the recovered volumetric geometry. We have evaluated the model using the ZJU-MoCap dataset, and the obtained highly competitive results demonstrate the effectiveness of our proposed Semantic-Human. We also showcase various compelling applications, including label denoising, label synthesis and image editing, and empirically validate its advantageous properties.
\end{abstract}

\section{Introduction}

\begin{figure}[t]
\centering
\includegraphics[width=\columnwidth]{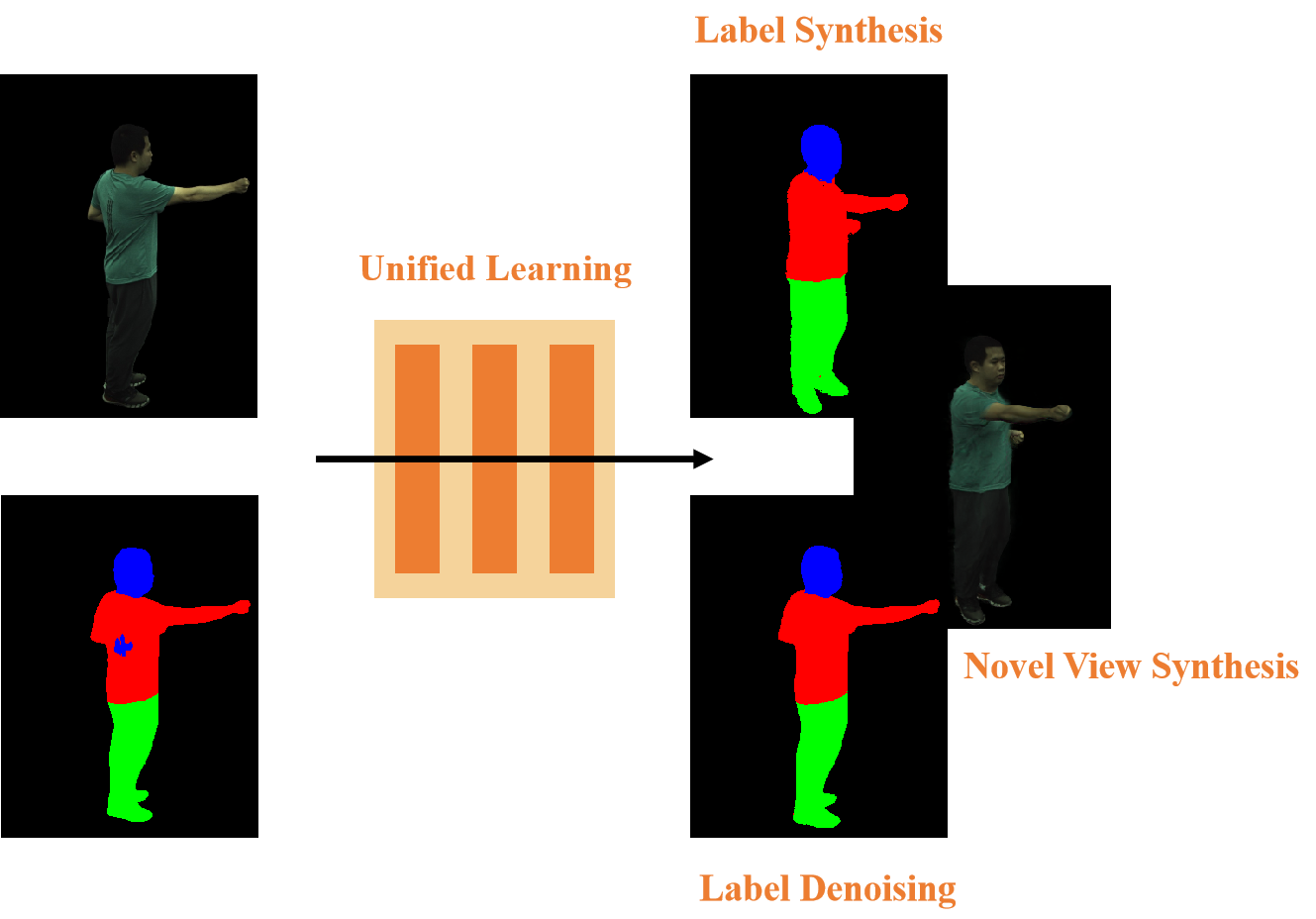} 
\caption{Neural radiance fields (NeRF) jointly encoding appearance and geometry contain strong priors for human parsing. We build upon this to design a unified learning framework for novel view synthesis, label denoising, label synthesis and image editing.}
\label{fig:motivation}
\end{figure}

Neural rendering of humans is a significant research direction in the field of computer vision due to its promising application scenarios. Previous methods for free-viewpoint rendering of humans have primarily focused on achieving photorealistic results without incorporating semantic labels. However, semantic information is crucial for human behavior analysis and image editing. Hence, achieving photorealistic free-viewpoint rendering of moving humans with human parsing from a monocular video remains an unsolved problem, especially with sparse and noisy semantic labels.

Neural Radiance Fields (NeRF) \citep{DBLP:journals/corr/abs-2003-08934} have greatly enhanced the quality of novel view synthesis, leading to a series of highly influential works. NeRF based Human-specific rendering have recently become popular for learning models using multi-view video \citep{peng2021neural} and monocular video and achieve impressive photorealistic results guided by SMPL \citep{loper2015smpl}. However, there are still some challenges that have not been effectively addressed at the moment. One major challenge is the significant error involved in motion fields from observation space to canonical space. In each frame, a 3D point in observation space was transformed into the canonical space by following the rigid transformation of its closest point on the estimated SMPL mesh \citep{jiang2022neuman}. This transformation becomes less accurate when the 3D point is far from the SMPL mesh. HumanNeRF solves for blend weight defined in canonical space ignoring the transformation of SMPL surface. In contrast to pervious methods, We focus on rigid transformations in the entire canonical space as HumanNeRF while introduced surface constraints to ensuring the accuracy of the transformation of the SMPL surface. In addition, we have observed that HumanNeRF exhibits issues of blurry and inconsistent boundaries in novel view synthesis. We attribute these problems to overfitting and, therefore, propose the inclusion of silhouette constraints to achieve clearer and more coherent boundaries. We demonstrate that this constraints provides significantly better results compared to existing approaches.

Currently, neural rendering methods for humans have not taken into account human parsing. In order to obtain human parsing results for an image, we can employ off-the-shelf models. However, these off-the-shelf human parsing methods usually do not provide segmentation results with viewpoint consistency. Human parsing means attaching semantic labels to human parts, which is strongly related to the geometry and appearance. Therefore, we simultaneously reconstruct the neural radiance field  and the neural segmentation field for several applications as shown in Figure \ref{fig:motivation}.

To this end, we introduce Semantic-Human, a novel approach for jointly learning appearance and human parsing within the NeRF framework for modeling human motions. Additionally, we introduce a novel surface loss in our framework. The surface loss aims to regularize the rigid deformation fields by minimizing the distance between the transformed SMPL vertices and the ground truth. Furthermore, we incorporate a silhouette loss based on binary segmentation masks to further regularize the training process. To achieve human parsing, we incorporate a multi-class cross-entropy loss to learn the neural segmentation field. Our approach enables photorealistic rendering of a moving human with human parsing, even in scenarios where sparse and noisy semantic labels are available.

In summary, our work makes the following contributions:

\begin{itemize}
\item We propose a novel method for neural rendering of human from monocular video, which generates accurate and consistent human parsing while maintaining photorealistic details.
\item We introduces surface constraints to the motion field from SMPL prior and silhouette regularization to the recovered volumetric geometry.
\item We extend NeRF to jointly encode semantics with appearance and geometry for human parsing even with noisy pseudo-label semantic supervision.
\item Our method achieves highly competitive results on the ZJU-MoCap dataset, demonstrating superior photorealistic rendering and robust human parsing with viewpoint consistency.
\end{itemize}
\begin{figure*}[htbp]
\centering
\includegraphics[width=2\columnwidth]{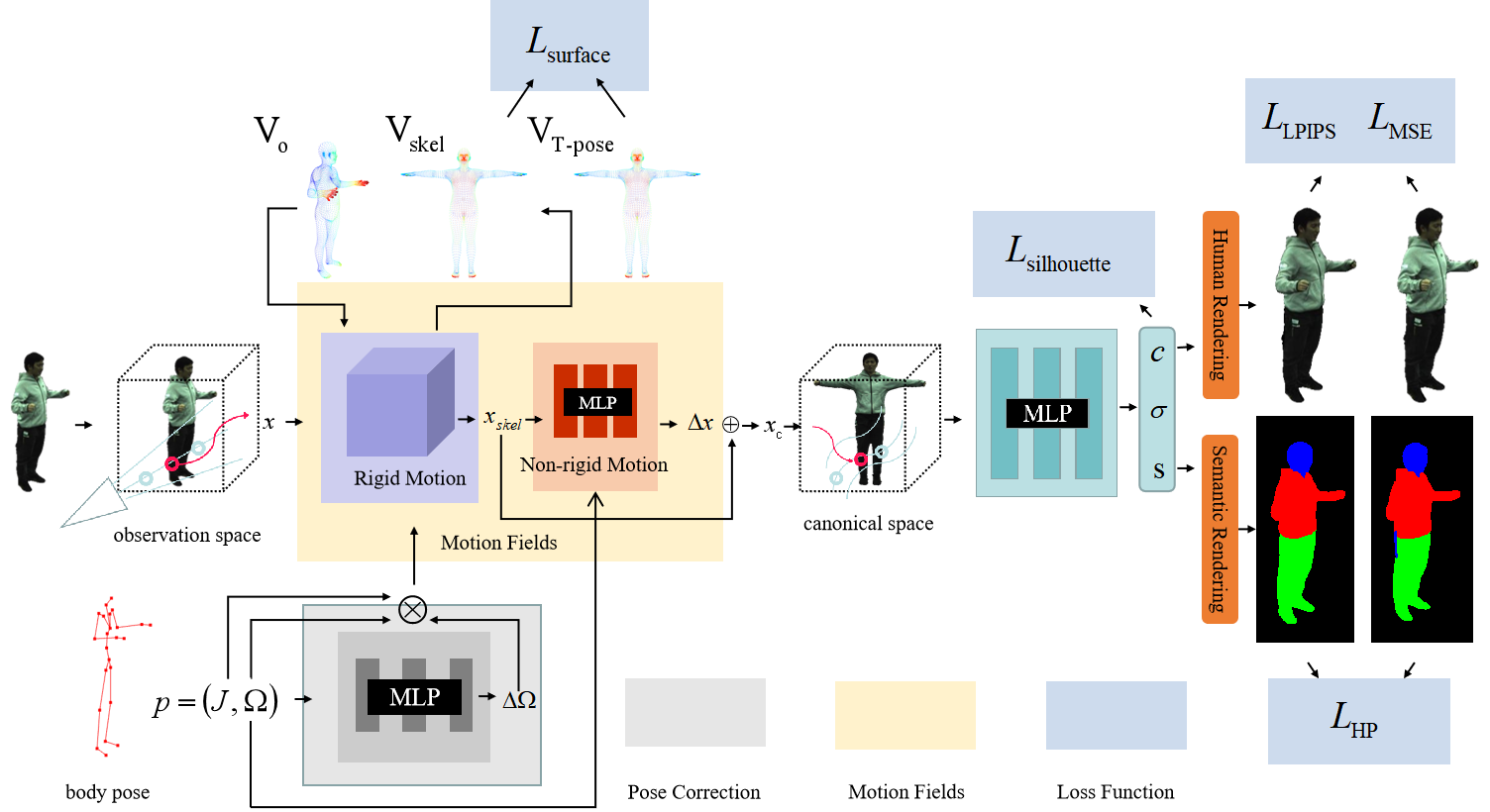} 
\caption{Given a video frame and its human parsing image as input, our method optimizes for canonical neural radiance fields and neural semantic fields. We will provide a detailed description of our method in the method section.}
\label{fig:overview}
\end{figure*}

\section{Related Work}
\subsection{Neural Rendering}
NeRF learns 3D scene representation using Multilayer Perceptron (MLP), which establishes a mapping from 3D coordinates and the viewing direction to corresponding values of color and density. NeRF demonstrate significant advantages in novel view synthesis and high-quality rendering \citep{barron2021mip,hedman2021baking,niemeyer2021giraffe,srinivasan2021nerv,zhang2020nerf++,zhang2021nerfactor}. Some recent works have attempted to combine NeRF with point clouds \citep{xu2022point,hu2023point2pix,hu2023trivol} and voxel grids \citep{liu2020neural,deng2023compressing}. The expansion of NeRF into dynamic scenes has yielded remarkable results \citep{liu2023robust,cao2023hexplane,fridovich2023k}. However, these methods primarily focus on scenes with limited motion.

\subsection{Neural Rendering of Humans}

Neural Rendering of Humans has emerged as a highly promising research direction. Previous works model human appearance and geometry as radiance fields under a multi-camera setting \citep{peng2021neural,peng2021animatable,shao2022doublefield,zheng2022structured,cheng2022generalizable,liu2021neural,kwon2021neural,zhao2022humannerf,chen2023uv}. Neural Body introduced for the first time the utilization of the SMPL for reconstructing NeRF of human \citep{peng2021neural}. To remove the multi-view constraints, some methods explore the reconstruction of humans from monocular videos \citep{weng2020vid2actor,jiang2022neuman,weng2022humannerf,jayasundara2023flexnerf,yu2023monohuman} and unstructured photos \citep{weng2023personnerf}. In HumanNeRF, the motion field is decoupled into two components: skeletal rigid field utilizing discrete grid weights and non-rigid deformation field utilizing MLP. The skeletal rigid field is responsible for handling coarse-grained deformations, while the non-rigid deformation field is designed to handle fine-grained deformations \citep{weng2022humannerf}.  MonoHuman introduces a consistent loss function to impose regularization on the motion field \citep{yu2023monohuman}. However, these methods are unable to generate semantic labels to specific human. The key distinction between our approach and previous studies lies in our utilization of noisy semantic labels as input. Building upon this, we have achieved high-fidelity human rendering and semantic rendering.

\subsection{Semantic Neural Rendering}
Semantic segmentation is important to scene understanding. Most semantic segmentation models require large collections of costly annotated data for training. Traditional 3D semantic segmentation methods often require expensive 3D annotated labels, such as point-based methods \citep{qi2017pointnet++}. Recently, some works have attempted to address the aforementioned issues by utilizing semantic neural rendering to achieve semantic segmentation. Semantic-NeRF is one of the pioneering works that extends NeRF in semantic segmentation. It incorporates semantics alongside appearance and geometry information in a joint manner \citep{zhi2021place}. NeSF trains a 3D semantic segmentation model with by utilizing a pre-trained NeRF model \citep{vora2021nesf}. CLA-NeRF learns a Category-Level articulated NeRF from rgb images with ground truth camera poses and part segments, which can perform view synthesis,part segmentation, and articulated pose estimation \citep{tseng2022cla}. Semantic Ray learns a generalizable semantic field with cross-reprojection attention \citep{liu2023semantic}. Sine propose a prior-guided editing field to encode fine-grained geometric and texture editing in 3D space and  enables users to edit a NeRF with image \citep{bao2023sine}. However, previous methods did not address human rendering with complex motions, achieving both photorealistic results and consistent segmentation results.

\subsection{Human Parseing}
Human parsing plays a pivotal role in the analysis of human behavior, with its primary objective being the segmentation of a human body into multiple anatomical regions \citep{gong2018instance}. Recent advanced algorithms in human synthesis with NeRF have incorporated binary segmentation constraints to eliminate the halo effects \citep{jayasundara2023flexnerf}. There has been limited research focus on the role of human parsing in human novel viewpoint generation. 
Our work is the first to investigate human rendering with complex motions, aiming to achieve both photorealistic results and consistent segmentation results.

\section{Method}
\label{sec:method}
\subsection{Preliminaries}
\subsubsection{HumanNeRF.}
HumanNeRF is the recent state-of-the-art (SOTA) approach \citep{weng2022humannerf}, represents a moving subject as a canonical volume $F_{c}$ warped to a body pose $\mathbf{p}$ to produce a volume $F_{o}$ in observed space. HumanNeRF establish a canonical space by utilizing the T-pose as a reference pose.
\begin{equation}
    F_{o}(\mathbf{x}, \mathbf{p})=F_{c}(T(\mathbf{x}, \mathbf{p}))
\end{equation}
The function $F_{c}$ maps 3D position to its corresponding color $\mathbf{c}$ and density $\sigma$ in the canonical space. Additionally, the motion field is defined by the function $T$ that maps points from the observed space ($\mathbf{x}_{o}$) back to the canonical space ($\mathbf{x}_{c}$). This mapping is guided by the observed pose $\mathbf{p}=(J, \Omega)$, where $J$ represents the standard 3D joint locations, and $\Omega=\left\{\boldsymbol{\omega}_{i}\right\}$ denotes the local joint rotations represented as axis-angle vectors $\boldsymbol{\omega}_{i}$.
The motion field is decomposed into two distinct components:
\begin{equation}
    T(\mathbf{x}, \mathbf{p})=T_{\text {skel }}(\mathbf{x}, \mathbf{p})+T_{\mathrm{NR}}\left(T_{\text {skel }}(\mathbf{x}, \mathbf{p}), \mathbf{p}\right),
\end{equation}
where $\mathbf{x}_{\text {skel }}=T_{\text {skel }}\left(\mathbf{x}, P_{\text {pose }}(\mathbf{p})\right)$, $T_{\mathrm{NR}}$ predicts a non-rigid offset $\Delta \mathbf{x}$, and $P_{\text {pose }}(\mathbf{p})$ corrects the body pose $\mathbf{p}=(J, \Omega)$ with the residual of joint angles $\Delta_{\Omega}$.
The skeletal motion $T_{\text {skel }}$ maps an observed position to the canonical space, computed as a weighted sum of $K$ motion bases $\left(R_{i}, \mathbf{t}_{i}\right):$
\begin{equation}
    T_{\text {skel }}(\mathbf{x}, \mathbf{p})=\sum_{i=1}^{K} w_{o}^{i}(\mathbf{x})\left(R_{i} \mathbf{x}+\mathbf{t}_{i}\right),
\end{equation}
where $\left(R_{i}, \mathbf{t}_{i}\right)$ explicitly computed from $\mathbf{p}$ indicates the rotation and translation that maps  $i \text {-th }$ bone from observation to canonical space and $w_{o}^{i}$ is the corresponding weight in observed space.

Each  $w_{o}^{i}$  is approximated using weights  $w_{c}^{i}$  defined in canonical space:
\begin{equation}
    w_{o}^{i}(\mathbf{x})=\frac{w_{c}^{i}\left(R_{i} \mathbf{x}+\mathbf{t}_{i}\right)}{\sum_{k=1}^{K} w_{c}^{k}\left(R_{k} \mathbf{x}+\mathbf{t}_{k}\right)} .
\end{equation}
HumanNeRF stores the set of $\left\{w_{c}^{i}(\mathbf{x})\right\}$ and a background class into a single volume grid $W_{c}(\mathbf{x})$ with $K+1$ channels, generated by a convolution network $\mathrm{CNN}_{\theta_{\text {skel }}}$ that takes as input a random (constant) latent code $\mathbf{z}$.


\begin{figure}[htbp]
\centering
\includegraphics[width=\columnwidth]{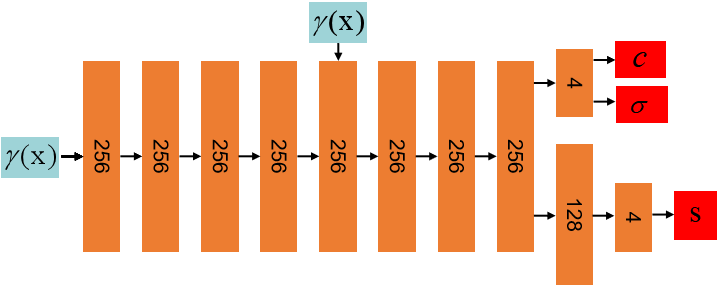} 
\caption{Semantic-Human neural fields network architecture. 3D position $\mathbf{x}$ are fed into the network after positional encoding (PE). Volume density $\sigma$, colours $\mathbf{c}$ and semantic logits $\mathbf{s}$ are functions of 3D position.}
\label{fig1}
\end{figure}

To render the observed volume $F_{o}$, which generates color $\mathbf{c}$ and density $\sigma$, HumanNeRF renders a neural field using the volume rendering equation as described by NeRF \citep{DBLP:journals/corr/abs-2003-08934}:
\begin{equation}
    C(\mathbf{r}) = \sum_{i = 1}^{D}\left(\prod_{j = 1}^{i-1}\left(1-\alpha_{j}\right)\right) \alpha_{i} \mathbf{c}\left(\mathbf{x}_{i}\right),
\end{equation}

\begin{equation}
    \alpha_{i}=f\left(\mathbf{x}_{i}\right)\left(1-\exp \left(-\sigma\left(\mathbf{x}_{i}\right) \Delta t_{i}\right)\right),
\end{equation}
where $f\left(\mathbf{x}\right)=\sum_{k=1}^{K} w_{c}^{k}\left(R_{k} \mathbf{x}+\mathbf{t}_{k}\right)$ is foreground likelihood.

\subsection{Semantic Neural Rendering}

As shown in Figure \ref{fig1}, we augment the NeRF by appending a segmentation head to jointly encode appearance, geometry and semantics. We formalize semantic segmentation as a function that maps a world coordinate $\mathbf{x}$ to a distribution over $L$ semantic labels via pre-softmax semantic logits $\mathbf{s}(\mathbf{x})$. The approximated expected semantic logits $S(\mathbf{r})$ of a ray $\mathbf{r}(t)=\mathbf{o}+t \mathbf{d}$ with $D$ samples is computed as:
\begin{equation}
    S(\mathbf{r}) = \sum_{i = 1}^{D}\left(\prod_{j = 1}^{i-1}\left(1-\alpha_{j}\right)\right) \alpha_{i} \mathbf{s}\left(\mathbf{x}_{i}\right).
\end{equation}

The semantic logits can be further processed by applying a softmax normalization layer, which transforms them into multi-class probabilities. 


\section{Semantic-Human Optimization}



\subsection{loss function}
While HumanNeRF performs well on monocular videos with complex motions, we have observed that it occasionally produces unsatisfactory results for novel viewpoints due to limited observations. We have identified two primary challenges in HumanNeRF: the limited generalization of its motion field to novel viewpoints and overfitting to the training set due to excessively strong non-rigid motion, leading to blurry and unsmooth boundaries. To address these limitations, we propose a straightforward approach by introducing two novel losses: the surface loss and silhouette loss. These losses are designed to constrain the motion field and neural radiance field, respectively.
\subsubsection{Surface Loss.}
To address the first limitation, we introduce the surface loss, as  illustrated in Figure \ref{fig:overview}. The ground truth is represented by SMPL vertices in T-pose (canonical space). By minimizing $\mathcal{L}_{\text {surface}}$, we constrain the skeletal rigid field,
\begin{equation}
    \mathcal{L}_{\text {surface}}=\|V_{\text {skel}}-V_{\text{T-pose}}\|_2^2,
\end{equation}
where $V_{\text {skel}}=T_{\text {skel }}(V_{\text {o}}, \mathbf{p})$, and $V_{\text {o}}$ are SMPL vertices in the observation space.
\subsubsection{Silhouette Loss.}

Given the binary human mask $\mathcal{M}$ for the observed image $I \in \mathcal{I}$, We address the second limitation by proposing a silhouette loss $\mathcal{L}_{\text {silhouette}}$ to get clean and shaper boundary,

\begin{equation}
    A(\mathbf{r}) = \sum_{i = 1}^{D}\left(\prod_{j = 1}^{i-1}\left(1-\alpha_{j}\right)\right) \alpha_{i},
\end{equation}

\begin{equation}
\hat{M}=\left\{\begin{array}{l}
1, \text { if } \mathcal{A}(r)>1 \\
\mathcal{A}(r), \text { otherwise },
\end{array}\right.
\end{equation}

\begin{equation}
    \mathcal{L}_{\text {silhouette}}=\|\hat{\mathcal{M}}-\mathcal{M}\|_2^2,
\end{equation}
where $A(\mathbf{r})$ is expected alpha mask. 


\subsubsection{Human Parsing Loss.}
The human parsing loss are computed for the set of all rays denoted as $R$, which is emitted from the query image $I \in \mathcal{I}$. The semantic loss is computed as multi-class crossentropy loss to encourage the rendered semantic labels to be consistent with the provided labels s, where $1 \leq l \leq L$ denotes the class index,
\begin{equation}
    \label{eq:Human Parsing}
    \mathcal{L}_{\text {HP}}=-\sum_{\mathbf{r} \in \mathcal{R}}\left(\sum_{l=0}^{L-1} p^{l}(\mathbf{r}) \log \hat{p}^{l}(\mathbf{r})\right),
\end{equation}
where $p^{l}$, $\hat{p}^{l}$ are the multi-class semantic probability at class $l$ of the ground truth map. The loss function $L_{HP}$ is selected as a multi-class cross-entropy loss in order to promote consistency between the rendered semantic labels and the provided labels, regardless of whether they are Pseudo-labels or sparse observations.

Additionally, we employ $\mathcal{L}_{\mathrm{LPIPS}}$ and $\mathcal{L}_{\mathrm{MSE}}$ as the reconstruction losses, where we select VGG as the backbone for LPIPS. Our final loss function is defined as follows:
\begin{equation}
\mathcal{L}=\lambda_{1} \mathcal{L}_{\mathrm{LPIPS}}+\lambda_{2} \mathcal{L}_{\mathrm{MSE}} + \lambda_{3} \mathcal{L}_{\text {silhouette}} + \lambda_{4} \mathcal{L}_{\text {silhouette}} + \lambda_{5} \mathcal{L}_{\text {HP}}.
\end{equation}

\subsection{Optimization Details}

\subsubsection{Delayed optimization of neural semantic field.}
We find that  both NeRF and neural semantic field have significant impacts on motion fields – parts of the human's motions is modeled by semantic neural rendering fields. Consequently, the quality of human rendering deteriorates due to imprecise motion fields caused by noisy semantic labels. To address this issue, we employ a delayed optimization process. We initiate the optimization process by disabling the human parsing loss and reintroduce the human parsing loss after 50K iters.
\begin{figure*}[htbp]
\centering
\resizebox{1.5\columnwidth}{!}{
\begin{tabular}{c|c|c|c}

\multicolumn{1}{c|}{} & \multicolumn{1}{c|}{Training View}              &\multicolumn{1}{c|}{} & \multicolumn{1}{c}{Training View}     \\ \cline{2-2}\cline{4-4}
Input & Ground Truth  \hspace{2cm}Ours\hspace{2cm}  Overlay &Input & Ground Truth  \hspace{2cm}Ours\hspace{2cm}  Overlay \\ \hline
\multicolumn{1}{c|}{\includegraphics[scale=0.32]{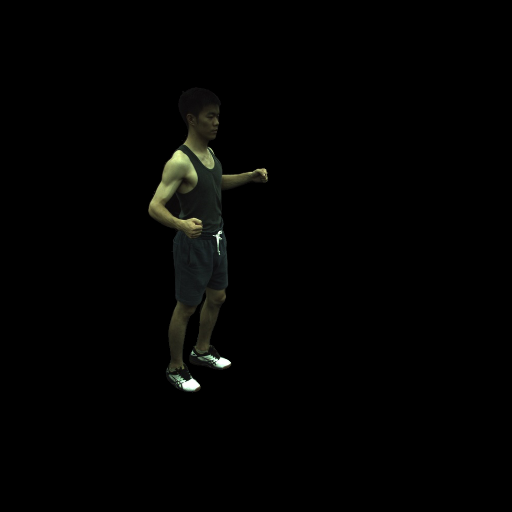}}          & \multicolumn{1}{c|}{\includegraphics[scale=0.4]{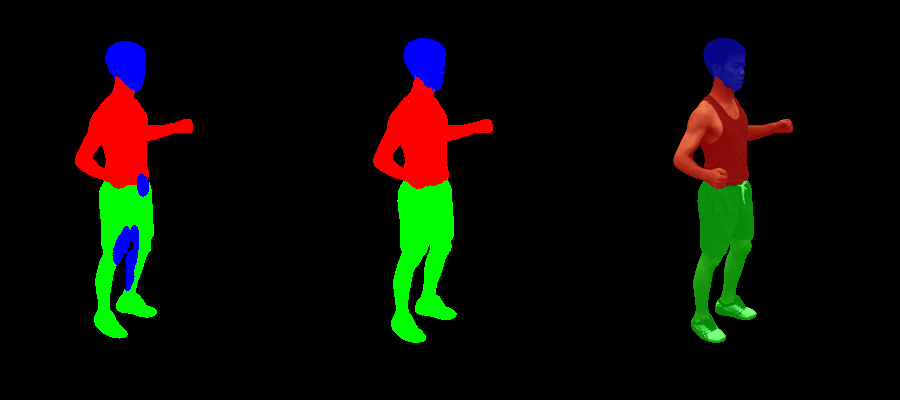}}       &\multicolumn{1}{c|}{\includegraphics[scale=0.32]{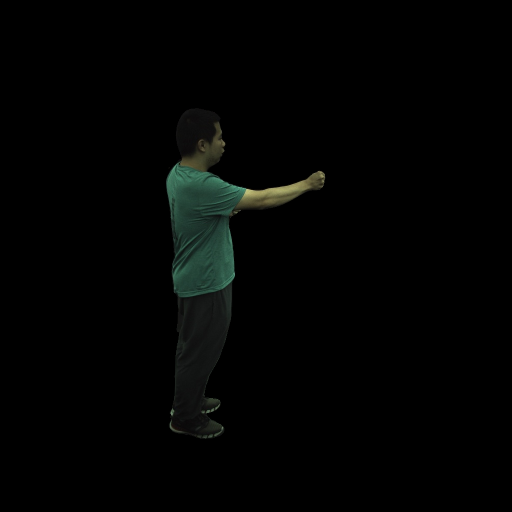}}                 & \multicolumn{1}{c}{\includegraphics[scale=0.4]{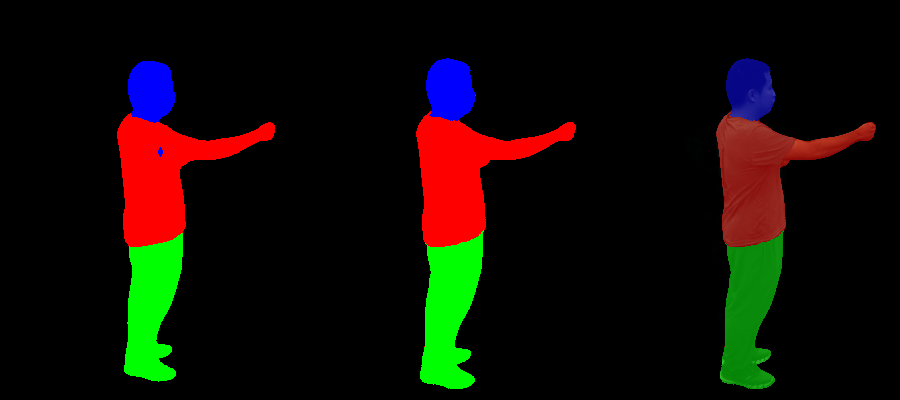}}                   \\ \hline
\multicolumn{1}{c|}{\includegraphics[scale=0.32]{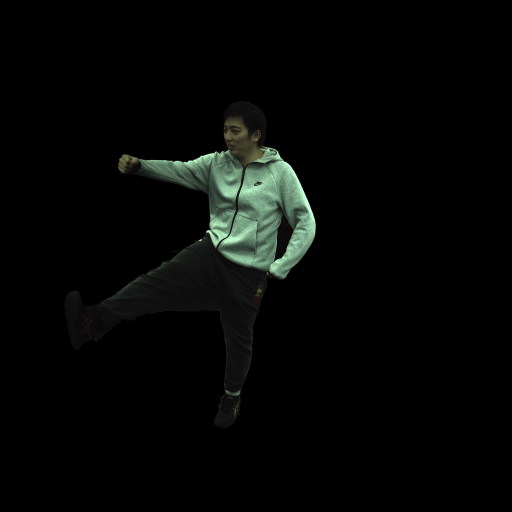}}           & \multicolumn{1}{c|}{\includegraphics[scale=0.4]{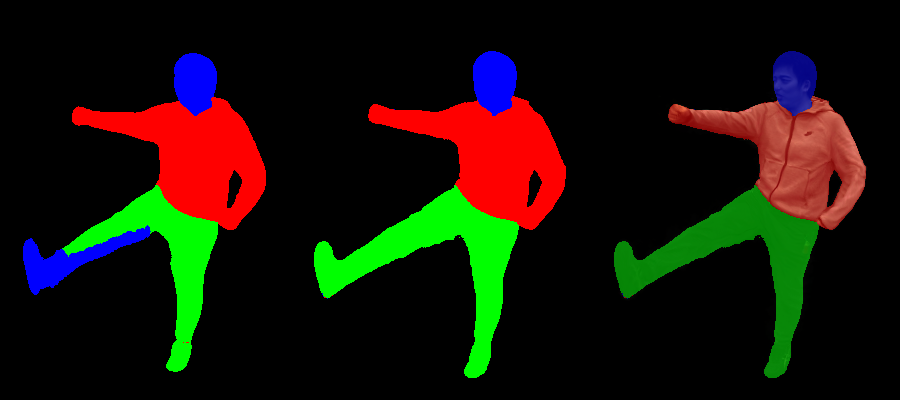}}      & \multicolumn{1}{c|}{\includegraphics[scale=0.32]{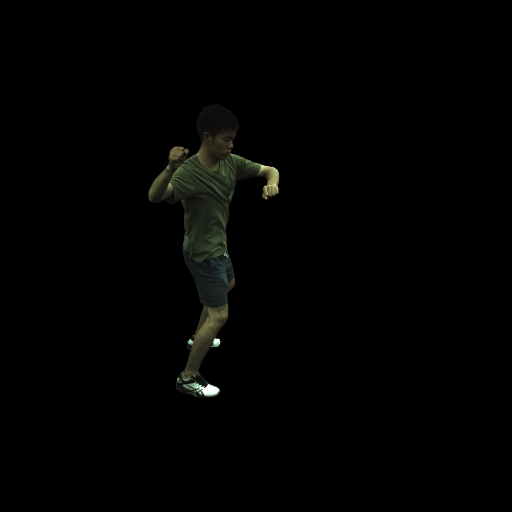}}                  & \multicolumn{1}{c}{\includegraphics[scale=0.4]{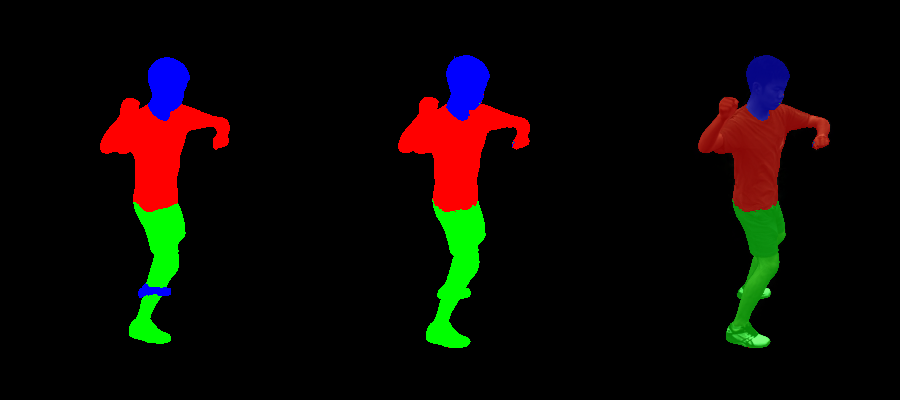}}          \\ \hline
\multicolumn{1}{c|}{\includegraphics[scale=0.32]{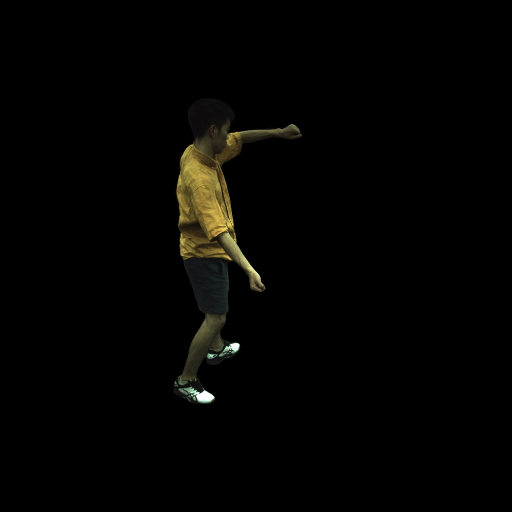}}           & \multicolumn{1}{c|}{\includegraphics[scale=0.4]{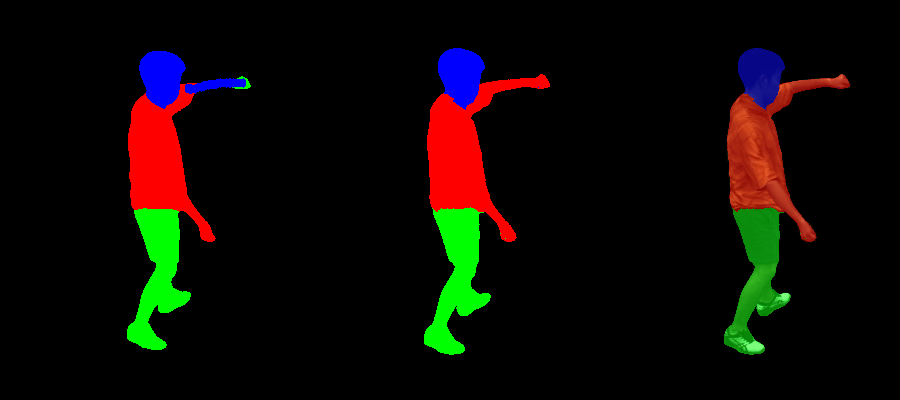}}      &\multicolumn{1}{c|}{\includegraphics[scale=0.32]{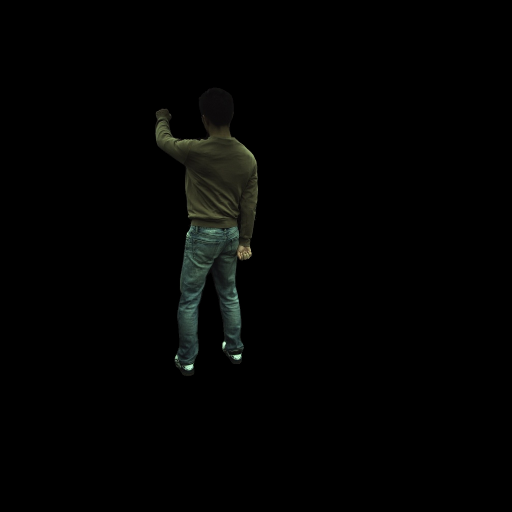}}                  & \multicolumn{1}{c}{\includegraphics[scale=0.4]{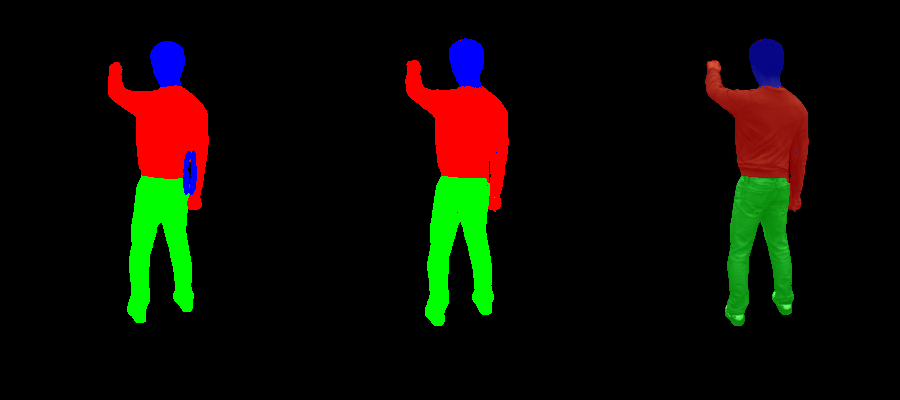}}          \\ \hline
\end{tabular}
}
\caption{Qualitative results of rendered human parsing on the ZJU-MoCap dataset. The off-the-shelf segmenter do not perform well in all scenarios, whereas our method allows for accurate human parsing. Zoom in for better view.}
\label{fig:SegComparison}
\end{figure*}

\begin{figure*}[htbp]
\centering
\resizebox{1.5\columnwidth}{!}{
\begin{tabular}{c|c|c|c}

\multicolumn{1}{c|}{} & \multicolumn{1}{c|}{Human Parsing}              &\multicolumn{1}{c|}{} & \multicolumn{1}{c}{Human Parsing}     \\ \cline{2-2}\cline{4-4}
Input & Ground Truth \hspace{3cm} Ours &Input & Ground Truth \hspace{3cm} Ours \\ \hline
\multicolumn{1}{c|}{\includegraphics[scale=0.32]{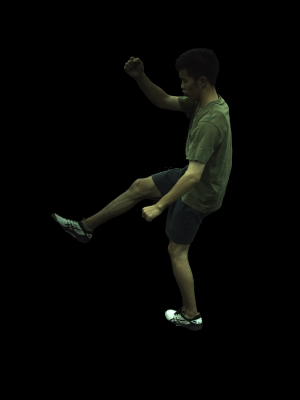}}          & \multicolumn{1}{c|}{\includegraphics[scale=0.33]{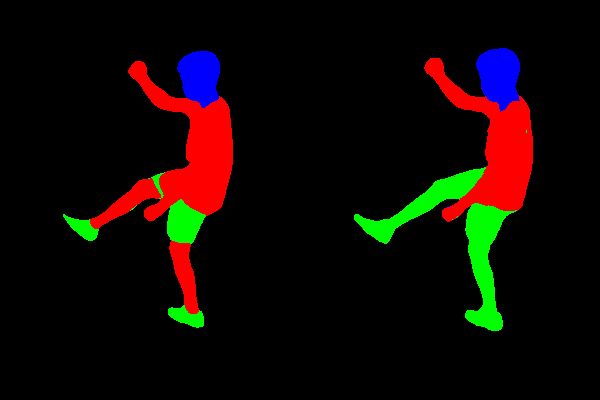}}       &\multicolumn{1}{c|}{\includegraphics[scale=0.32]{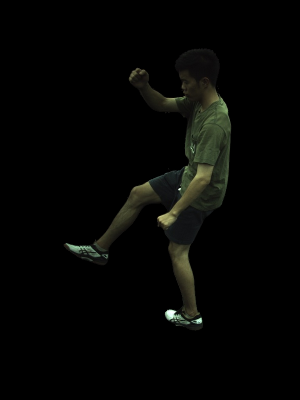}}                 & \multicolumn{1}{c}{\includegraphics[scale=0.35]{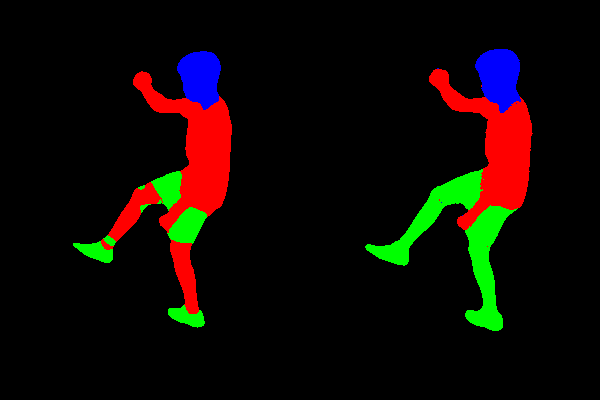}}                   \\ \hline
\multicolumn{1}{c|}{\includegraphics[scale=0.32]{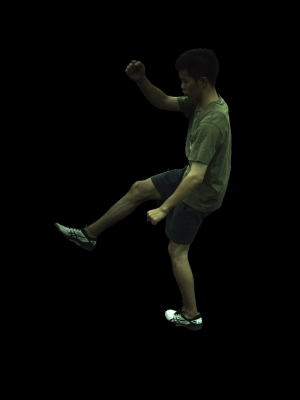}}           & \multicolumn{1}{c|}{\includegraphics[scale=0.33]{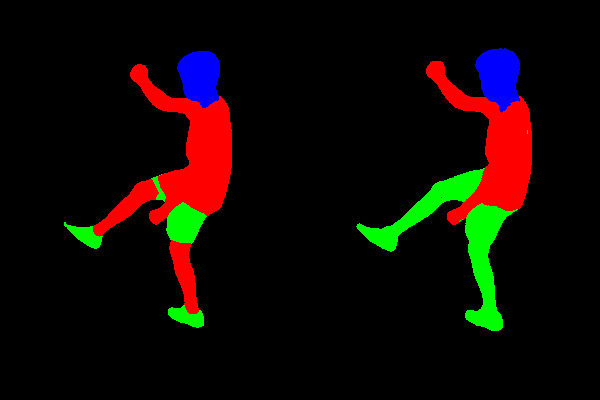}}      & \multicolumn{1}{c|}{\includegraphics[scale=0.32]{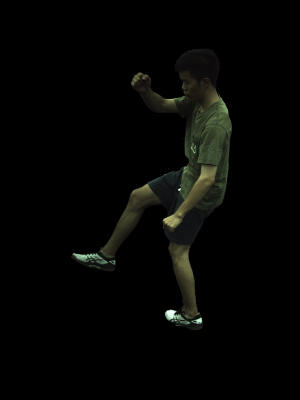}}                  & \multicolumn{1}{c}{\includegraphics[scale=0.35]{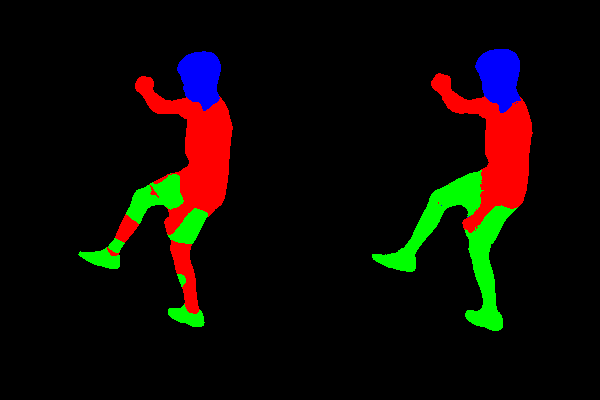}}          \\ \hline
\multicolumn{1}{c|}{\includegraphics[scale=0.32]{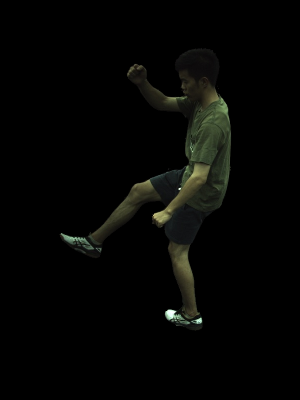}}           & \multicolumn{1}{c|}{\includegraphics[scale=0.33]{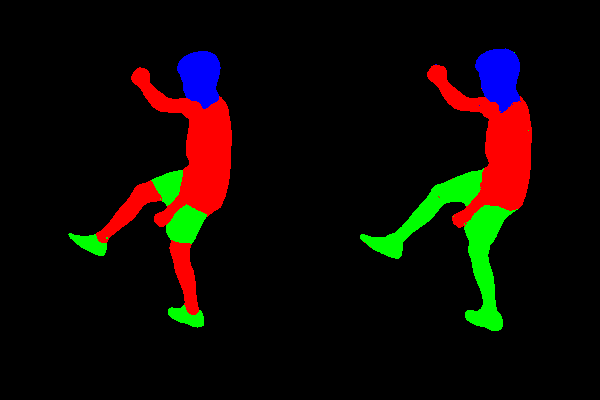}}      &\multicolumn{1}{c|}{\includegraphics[scale=0.32]{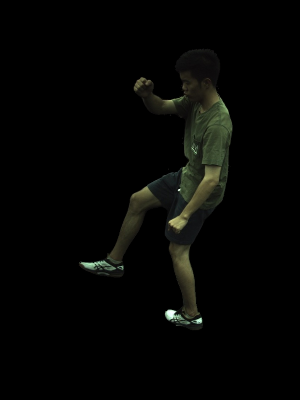}}                  & \multicolumn{1}{c}{\includegraphics[scale=0.35]{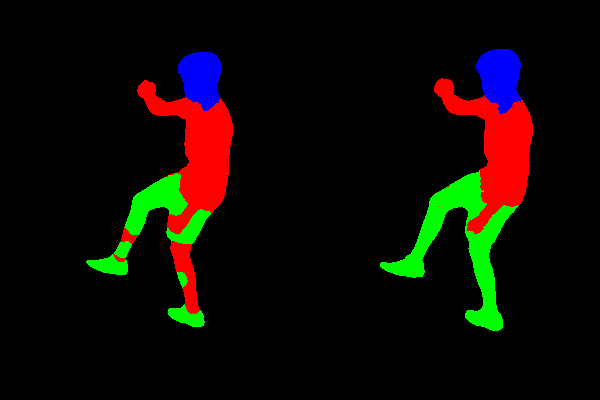}}          \\ \hline
\end{tabular}
}
\caption{Qualitative results of the rendered viewpoint-consistent human parsing on the ZJU-MoCap dataset. Given a sequence of consecutive frame images, the off-the-shelf segmenter cannot get viewpoint-consistent segmentation, whereas our method allows for consistent human parsing. Zoom in for better view.}
\label{fig:seg_consis}
\end{figure*}

\begin{figure*}[htbp]
\centering
\resizebox{1.5\columnwidth}{!}{
\begin{tabular}{c|ccc|ccc}
\multicolumn{1}{c|}{\multirow{2}{*}{Input}} & \multicolumn{3}{c|}{ Novel View 1}                                         & \multicolumn{3}{c}{ Novel View 2}                                         \\ \cline{2-7} 
\multicolumn{1}{c|}{}            & HumanNeRF &  \hspace{3cm}Ours\hspace{2cm}  & Ground Truth                & HumanNeRF & \hspace{3cm} Ours\hspace{2cm} & Ground Truth \\ \hline

\multicolumn{1}{c|}{\includegraphics[scale=0.35]{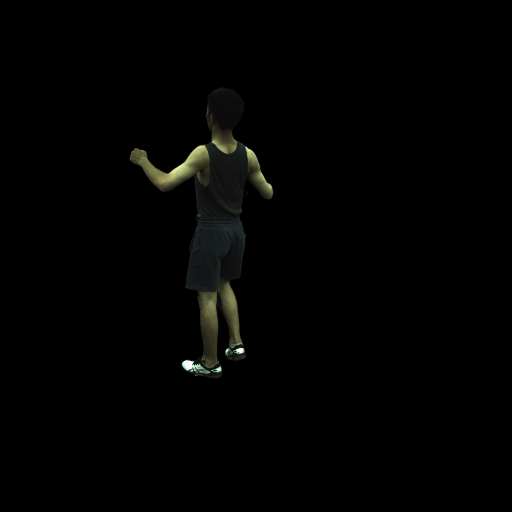}}                  & \multicolumn{3}{c|}{\includegraphics[scale=0.8]{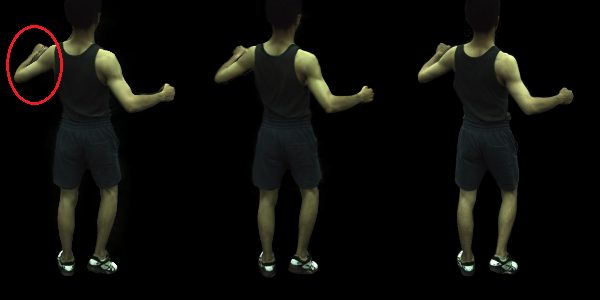}}                               & \multicolumn{3}{c}{\includegraphics[scale=0.8]{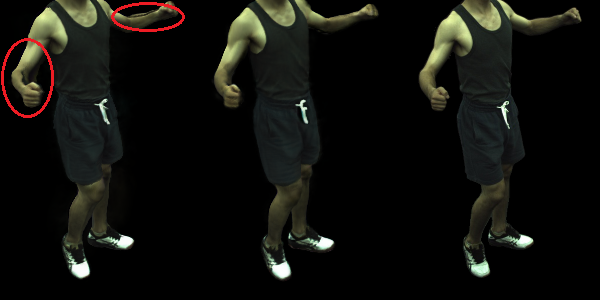}}         \\ \hline
\multicolumn{1}{c|}{\includegraphics[scale=0.35]{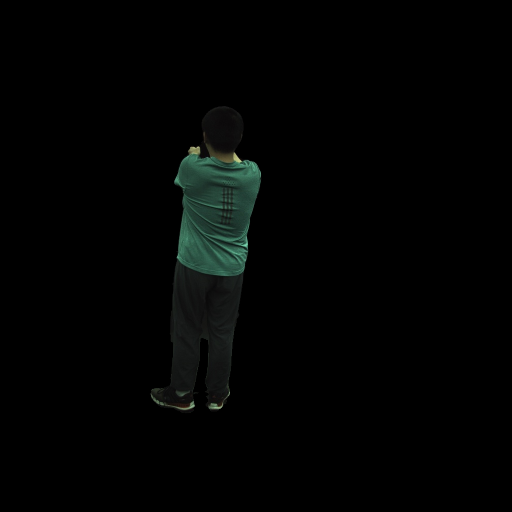}}                  & \multicolumn{3}{c|}{\includegraphics[scale=0.8]{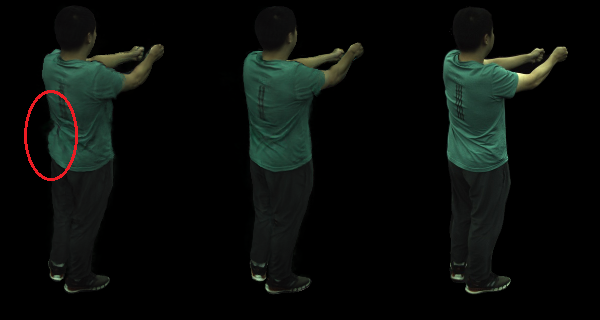}}                               & \multicolumn{3}{c}{\includegraphics[scale=0.8]{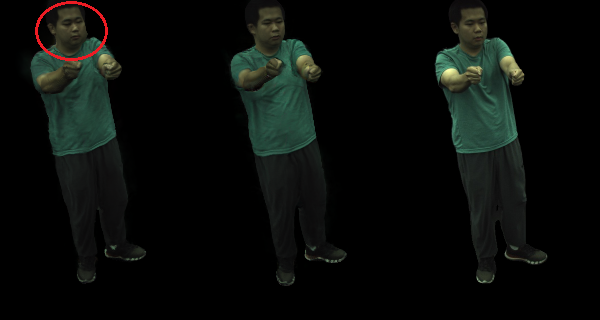}}         \\ \hline
\multicolumn{1}{c|}{\includegraphics[scale=0.35]{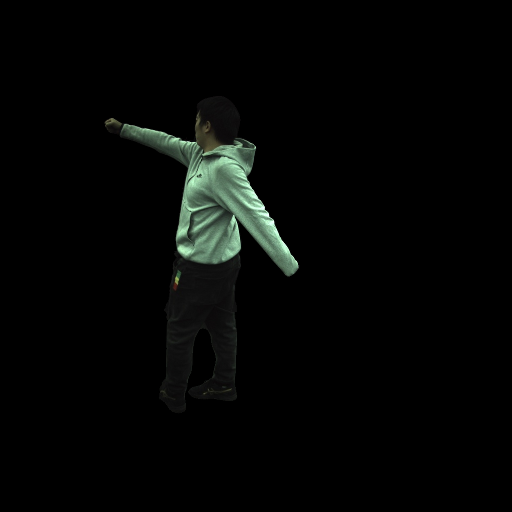}}                  & \multicolumn{3}{c|}{\includegraphics[scale=0.8]{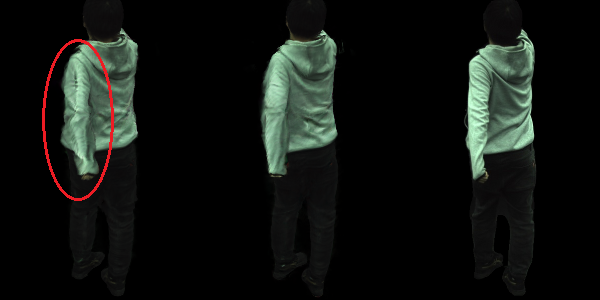}}                               & \multicolumn{3}{c}{\includegraphics[scale=0.8]{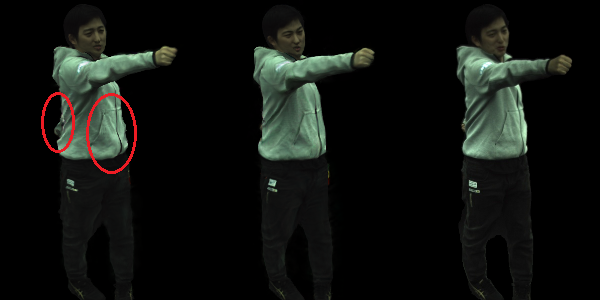}}          \\ \hline
\multicolumn{1}{c|}{\includegraphics[scale=0.35]{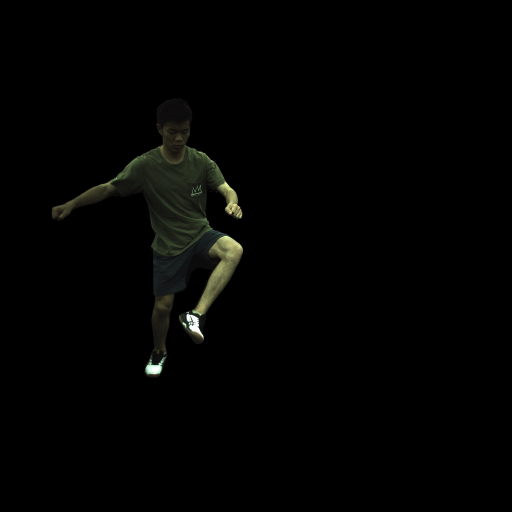}}                  & \multicolumn{3}{c|}{\includegraphics[scale=0.8]{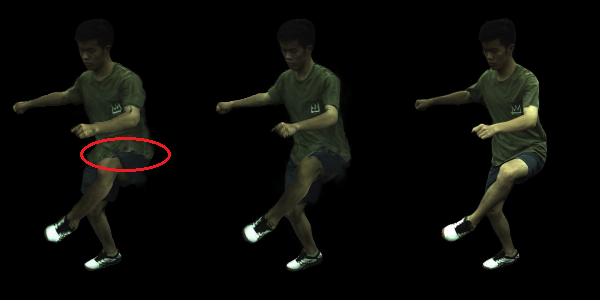}}                               & \multicolumn{3}{c}{\includegraphics[scale=0.8]{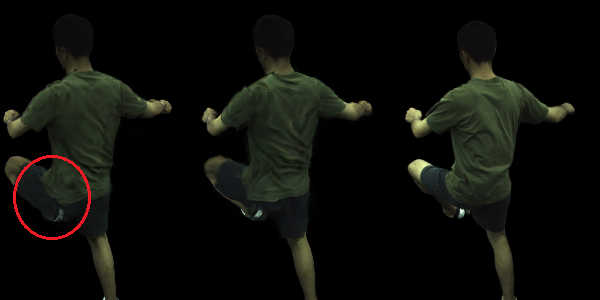}}          \\ \hline
\multicolumn{1}{c|}{\includegraphics[scale=0.35]{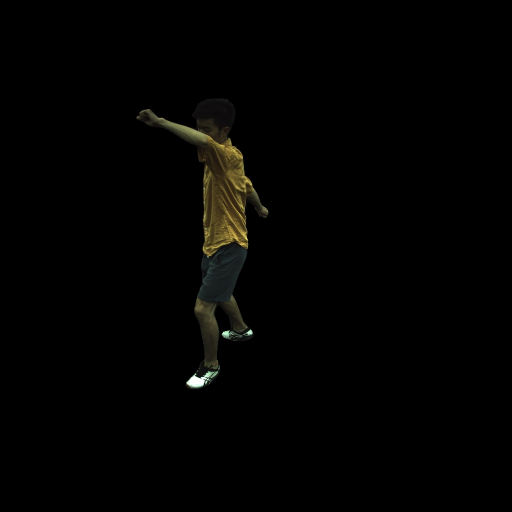}}                  & \multicolumn{3}{c|}{\includegraphics[scale=0.8]{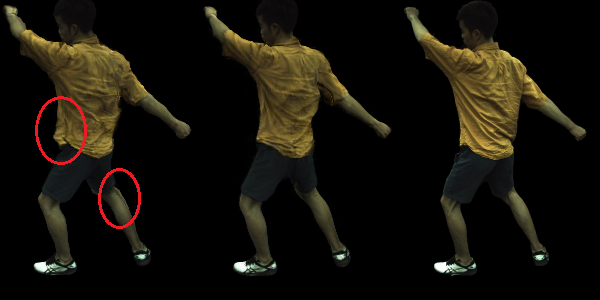}}                               & \multicolumn{3}{c}{\includegraphics[scale=0.8]{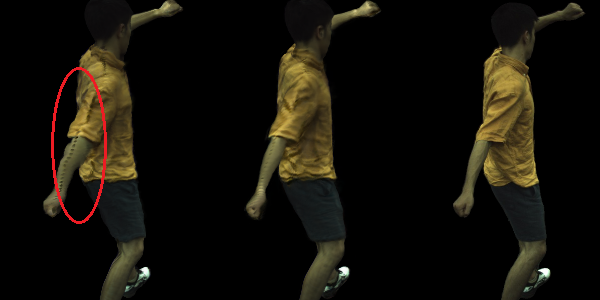}}           \\ \hline
\multicolumn{1}{c|}{\includegraphics[scale=0.35]{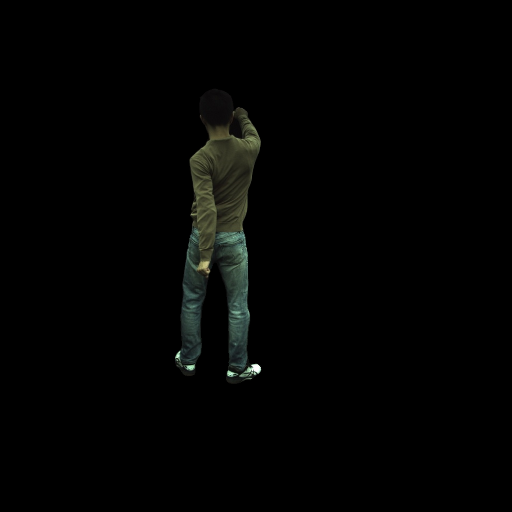}}                  & \multicolumn{3}{c|}{\includegraphics[scale=0.8]{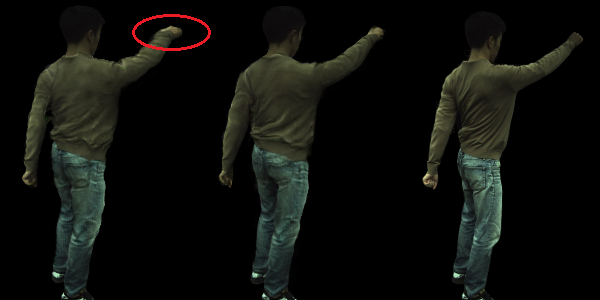}}                               & \multicolumn{3}{c}{\includegraphics[scale=0.8]{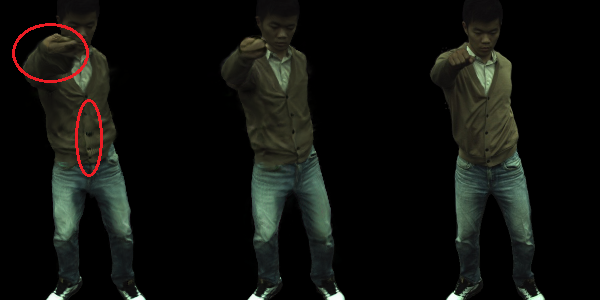}}            \\ \hline
\end{tabular}
}
\caption{Qualitative comparison of rendered novel views on the ZJU-MoCap dataset. Zoom in for better view.}
\label{fig:QualitativeComparison}
\end{figure*}

\begin{figure}[t]
\centering
\includegraphics[width=\columnwidth]{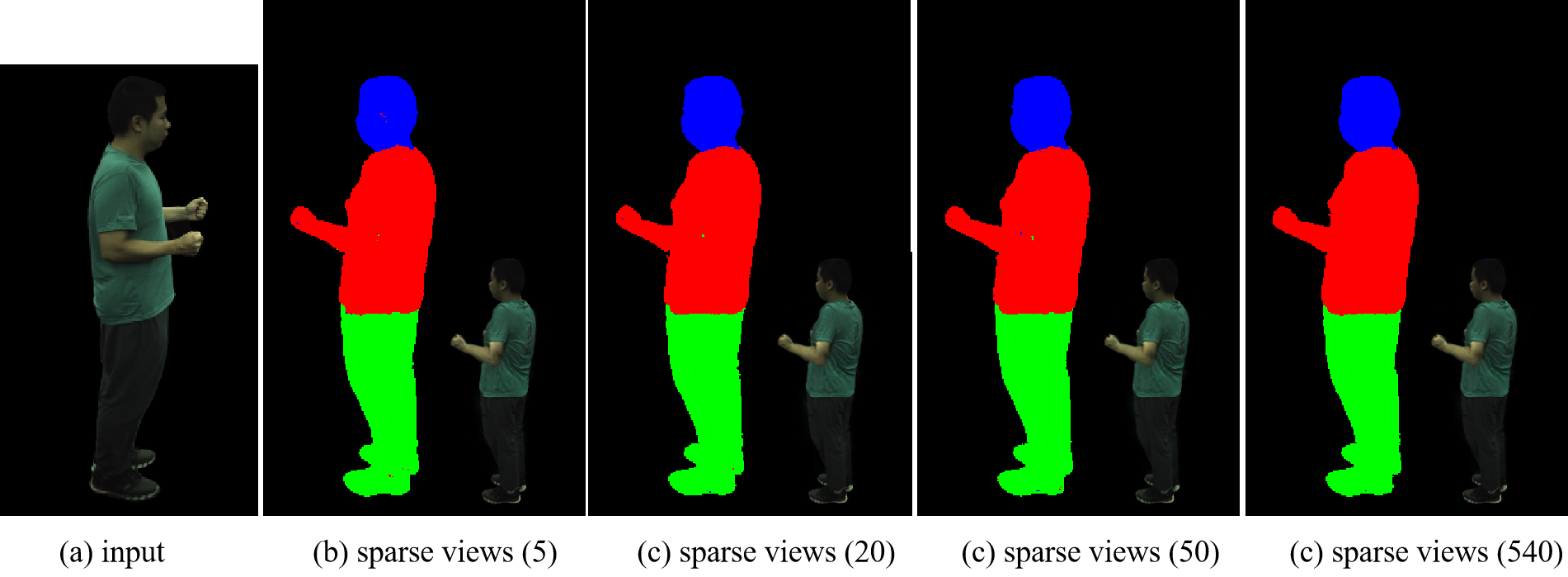} 
\caption{Novel view label synthesis of model trained with different number of pseudo-labels images. The model of (b) is trained with only 5 pseudo-labels, (c) 20 labels, (d) 50 labels, and (e) 540 labels (full). Zoom in for better view.}
\label{fig:abl-segnum}
\end{figure}

\begin{figure}[t]
\centering
\includegraphics[width=\columnwidth]{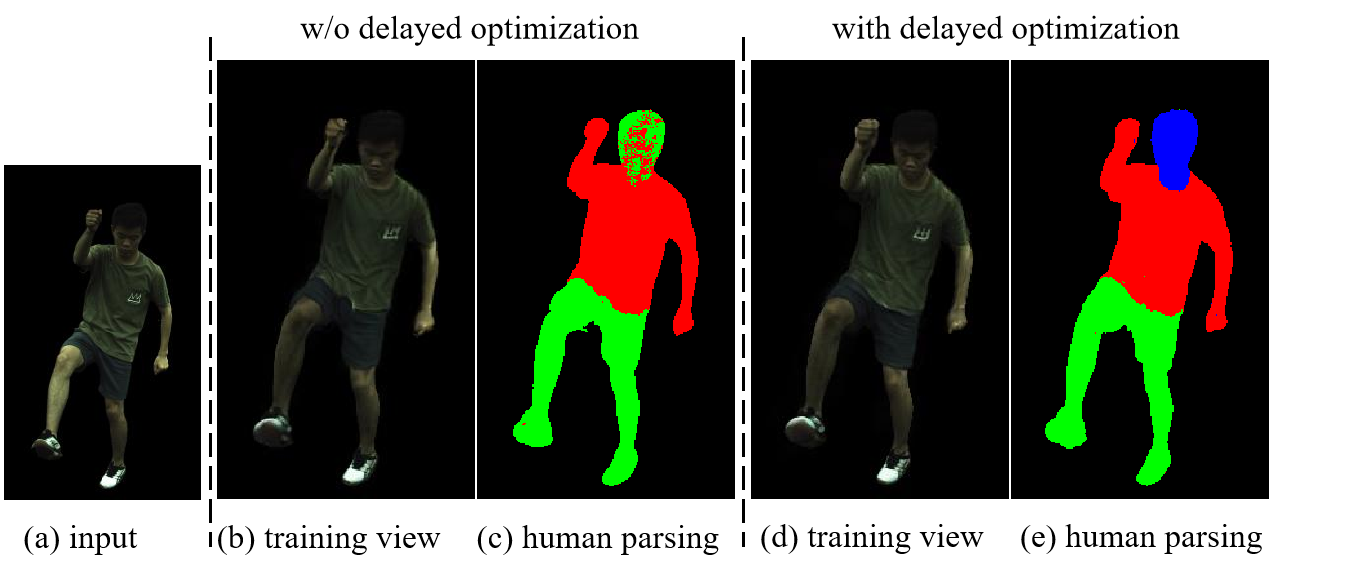} 
\caption{Delayed optimization (d, e) leads to better neural rendering than the result without it (b, c). Delayed optimization of human parsing allows the network to initially construct the motion field supervised by color images. Zoom in for better view.}
\label{fig:delay}
\end{figure}

\begin{figure}[t]
\centering
\includegraphics[width=0.8\columnwidth]{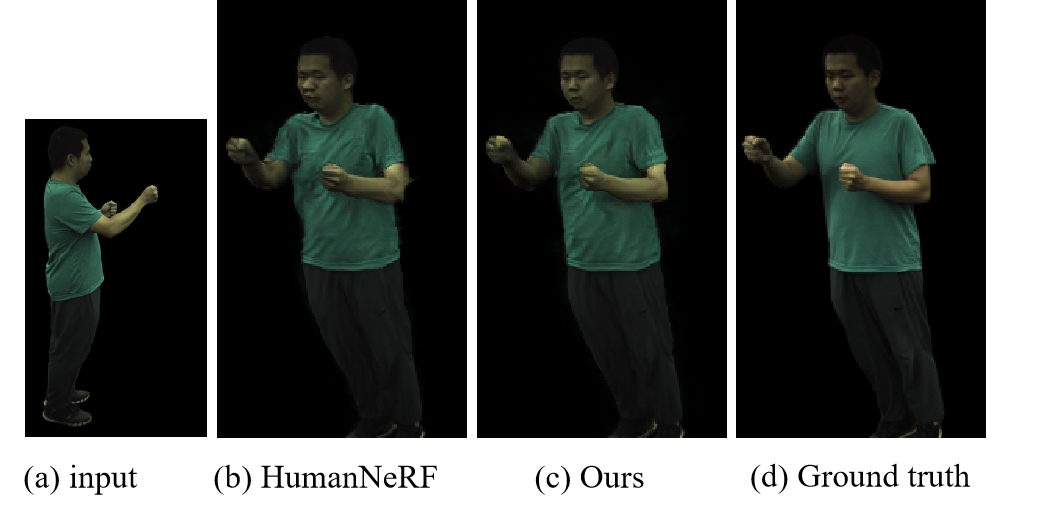} 
\caption{Surface loss (c) contributes to better generalization in novel view synthesis than the result without it (b), and can provide additional constraints from SMPL priors to the skeletal rigid field. Zoom in for better view.}
\label{fig:abl-surface}
\end{figure}


\begin{table*}[htbp]
\begin{tabular}{|c|c|c|c|c|c|c|c|c|c|}
\hline \multicolumn{1}{|c|}{} & \multicolumn{3}{c|}{ Subject 377 } & \multicolumn{3}{c|}{ Subject 386 } & \multicolumn{3}{c|}{ Subject 387 } \\
\cline { 2 - 10 } & PSNR $\uparrow$ & SSIM $\uparrow$ & LPIPS* $\downarrow$ & PSNR $\uparrow$ & SSIM $\uparrow$ & LPIPS* $\downarrow$ & PSNR $\uparrow$ & SSIM $\uparrow$ & LPIPS* $\downarrow$ \\
\hline HumanNeRF & 30.41 & 0.9743 & 24.06 & 33.20 & 0.9752 & 28.99 & 28.18 & \textbf{0.9632} & \textbf{35.58} \\
\hline Ours & \textbf{30.42} & \textbf{0.9769} & \textbf{22.73} & \textbf{33.43} & \textbf{0.9760} & \textbf{28.36} & \textbf{28.22} & 0.9631 & 36.94 \\

\hline \multicolumn{1}{|c|}{} & \multicolumn{3}{c|}{ Subject 392 } & \multicolumn{3}{c|}{ Subject 393 } & \multicolumn{3}{c|}{ Subject 394 } \\
\cline { 2 - 10 } & PSNR $\uparrow$ & SSIM $\uparrow$ & LPIPS* $\downarrow$ & PSNR $\uparrow$ & SSIM $\uparrow$ & LPIPS* $\downarrow$ & PSNR $\uparrow$ & SSIM $\uparrow$ & LPIPS* $\downarrow$ \\
\hline  HumanNeRF & \textbf{31.04} & \textbf{0.9705} & \textbf{32.12} & 28.31 & 0.9603 & \textbf{36.72} & \textbf{30.31} & \textbf{0.9642} & \textbf{32.89} \\
\hline Ours & 30.80 & 0.9696 & 33.54 & \textbf{28.52} & \textbf{0.9610} & 37.25 & 30.08 & 0.9630 & 34.14 \\
\hline
\end{tabular}
\caption{Quantitative comparison on ZJU-MoCap dataset. LPIPS* $=$ LPIPS $\times 10^3$.}
\label{tab:comparison on zjumocap}
\end{table*}

 \begin{table*}[htbp]
\centering
\begin{tabular}{|c|c|c|c|c|c|c|}
\hline Ablation & $\mathbf{P S N R} \uparrow$ & $\Delta$ & $\mathbf{S S I M} \uparrow$ & $\Delta \times \mathbf{1 0}^{-\mathbf{2}}$ & $\mathbf{L P I P S} \times \mathbf{1 0}^{\mathbf{3}} \downarrow$ & $\Delta$ \\
\hline HumanNeRF & 30.24 & 0.00 & 0.9679 & 0.00 & 31.73 & 0.00 \\
\hline Ours (full) & 30.25 & +0.01 & 0.9683 & +0.04 & 32.16 & +0.43 \\
\hline Ours (w/o HPL) & 30.22 & -0.02 & 0.9683 & +0.04 & 31.87 & +0.14 \\
\hline Ours (w/o silhouette loss) & \textbf{30.27} &  \textbf{+0.03} & 0.9668 &  -0.11 & \textbf{30.63} & \textbf{-1.1}  \\
\hline Ours (w/o surface loss) & 30.26  &  +0.02 &  \textbf{0.9685} &  \textbf{+0.06} & 31.74  &  +0.01 \\
\hline
\end{tabular}
\caption{Ablation: Effect of removing various modules and losses from our full approach on the ZJU-MoCap dataset. HPL: Human Parsing Loss.}
\label{tab:ablation on zjumocap}
\end{table*}

\section{Experiments}

\subsection{Evaluation method}
We compare our method with HumanNeRF \citep{weng2022humannerf} on ZJU-MoCap dataset \citep{peng2021neural}. The main differences  lie in we have achieved photorealistic rendering and human parsing and introduce a prior on SMPL vertices.

\subsection{Results and Analysis}
As shown in Table \ref{tab:comparison on zjumocap}, Semantic-Human outperforms HumanNeRF for subjects 377 and 386 under all metrics. Although HumanNeRF can produce high-quality rendered results, it still suffers from some artifacts such as unclear body contours and indistinct body part boundaries. Figure \ref{fig:QualitativeComparison}  shows that Semantic-Human's visual quality is better than HumanNeRF for ZJU-MoCap dataset.

\subsubsection{Qualitative Results of Rendered Training Views.}

As depicted in Figure \ref{fig:SegComparison}, our method is capable of generating results that are more accurate and reasonable than the ground truth, effectively reducing noise in the semantic labels.

\subsubsection{Qualitative Results with Semantic Consistency.}

Taking six consecutive frames as an example, we can observe that the pseudo labels contain a significant amount of noise and the semantic labels are highly inconsistent. In contrast, our proposed method is capable of reducing the noise in the semantic labels while generating consistent semantic labels across different frames as shown in Figure \ref{fig:seg_consis}.

\subsubsection{Analysis of Number of Semantic Supervision Views.}

We found that the more semantic supervision, the better the results of human parsing. Additionally, we discovered that even with supervision from just 5 human parsing images, surprising results can be generated. It can be observed that the quality of novel view label synthesis varies under different supervision conditions in Figure \ref{fig:abl-segnum}.

\subsection{Ablation studies}
Table \ref{tab:ablation on zjumocap} shows the effect of removing various modules and loss from our full approach. We observe that not all the proposed losses contribute to the performance improvement in numerical metrics. We indeed have some evidence suggesting improvements in rendering results in Figure \ref{fig:QualitativeComparison}. From Figure \ref{fig:abl-surface}, it is evident that our approach, after incorporating surface constraints, effectively constrains the rendering of the human body surface. Figure \ref{fig:delay} shows the utilization of the delayed optimization strategy leads to a notable enhancement in the outcomes of neural rendering of human and human parsing, especially for human parsing. Without delayed optimization, noisy human parsing supervision tends to blur the motion field and neural rendering of humans.
 
\section{Discussion}
\subsection{Limitations}
While our method showcases remarkable performance, it is crucial to acknowledge certain limitations. Specifically, our methodology has not been assessed in more refined human parsing experiments that involve accurate semantic labels. Furthermore, neural rendering of human conducted in the setting of sparse color image and sparse semantic supervision. These limitations indicate various interesting directions for future research.
\subsection{Conclusion}
We propose Semantic-Human, a novel approach for rendering human subjects with human parsing from monocular video. Our method achieves viewpoint consistency in human parsing while maintaining high-quality rendering results  by extending the NeRF. We hope the result points in a promising direction toward achieving both photorealistic rendering and accurate semantic rendering of human.
\bibliography{aaai23}

\begin{thebibliography}{38}
\providecommand{\natexlab}[1]{#1}

\bibitem[{Bao et~al.(2023)Bao, Zhang, Yang, Fan, Yang, Bao, Zhang, and
  Cui}]{bao2023sine}
Bao, C.; Zhang, Y.; Yang, B.; Fan, T.; Yang, Z.; Bao, H.; Zhang, G.; and Cui,
  Z. 2023.
\newblock Sine: Semantic-driven image-based nerf editing with prior-guided
  editing field.
\newblock In \emph{Proceedings of the IEEE/CVF Conference on Computer Vision
  and Pattern Recognition}, 20919--20929.

\bibitem[{Barron et~al.(2021)Barron, Mildenhall, Tancik, Hedman,
  Martin-Brualla, and Srinivasan}]{barron2021mip}
Barron, J.~T.; Mildenhall, B.; Tancik, M.; Hedman, P.; Martin-Brualla, R.; and
  Srinivasan, P.~P. 2021.
\newblock Mip-nerf: A multiscale representation for anti-aliasing neural
  radiance fields.
\newblock In \emph{Proceedings of the IEEE/CVF International Conference on
  Computer Vision}, 5855--5864.

\bibitem[{Cao and Johnson(2023)}]{cao2023hexplane}
Cao, A.; and Johnson, J. 2023.
\newblock Hexplane: A fast representation for dynamic scenes.
\newblock In \emph{Proceedings of the IEEE/CVF Conference on Computer Vision
  and Pattern Recognition}, 130--141.

\bibitem[{Chen et~al.(2023)Chen, Wang, Chen, Zhang, Li, Guo, Wang, and
  Wang}]{chen2023uv}
Chen, Y.; Wang, X.; Chen, X.; Zhang, Q.; Li, X.; Guo, Y.; Wang, J.; and Wang,
  F. 2023.
\newblock UV Volumes for real-time rendering of editable free-view human
  performance.
\newblock In \emph{Proceedings of the IEEE/CVF Conference on Computer Vision
  and Pattern Recognition}, 16621--16631.

\bibitem[{Cheng et~al.(2022)Cheng, Xu, Piao, Qian, Wu, Lin, and
  Li}]{cheng2022generalizable}
Cheng, W.; Xu, S.; Piao, J.; Qian, C.; Wu, W.; Lin, K.-Y.; and Li, H. 2022.
\newblock Generalizable neural performer: Learning robust radiance fields for
  human novel view synthesis.
\newblock \emph{arXiv preprint arXiv:2204.11798}.

\bibitem[{Deng and Tartaglione(2023)}]{deng2023compressing}
Deng, C.~L.; and Tartaglione, E. 2023.
\newblock Compressing explicit voxel grid representations: fast nerfs become
  also small.
\newblock In \emph{Proceedings of the IEEE/CVF Winter Conference on
  Applications of Computer Vision}, 1236--1245.

\bibitem[{Fridovich-Keil et~al.(2023)Fridovich-Keil, Meanti, Warburg, Recht,
  and Kanazawa}]{fridovich2023k}
Fridovich-Keil, S.; Meanti, G.; Warburg, F.~R.; Recht, B.; and Kanazawa, A.
  2023.
\newblock K-planes: Explicit radiance fields in space, time, and appearance.
\newblock In \emph{Proceedings of the IEEE/CVF Conference on Computer Vision
  and Pattern Recognition}, 12479--12488.

\bibitem[{Gong et~al.(2018)Gong, Liang, Li, Chen, Yang, and
  Lin}]{gong2018instance}
Gong, K.; Liang, X.; Li, Y.; Chen, Y.; Yang, M.; and Lin, L. 2018.
\newblock Instance-level human parsing via part grouping network.
\newblock In \emph{Proceedings of the European conference on computer vision
  (ECCV)}, 770--785.

\bibitem[{Hedman et~al.(2021)Hedman, Srinivasan, Mildenhall, Barron, and
  Debevec}]{hedman2021baking}
Hedman, P.; Srinivasan, P.~P.; Mildenhall, B.; Barron, J.~T.; and Debevec, P.
  2021.
\newblock Baking neural radiance fields for real-time view synthesis.
\newblock In \emph{Proceedings of the IEEE/CVF International Conference on
  Computer Vision}, 5875--5884.

\bibitem[{Hu et~al.(2023{\natexlab{a}})Hu, Xu, Chu, and Jia}]{hu2023trivol}
Hu, T.; Xu, X.; Chu, R.; and Jia, J. 2023{\natexlab{a}}.
\newblock TriVol: Point Cloud Rendering via Triple Volumes.
\newblock In \emph{Proceedings of the IEEE/CVF Conference on Computer Vision
  and Pattern Recognition}, 20732--20741.

\bibitem[{Hu et~al.(2023{\natexlab{b}})Hu, Xu, Liu, and Jia}]{hu2023point2pix}
Hu, T.; Xu, X.; Liu, S.; and Jia, J. 2023{\natexlab{b}}.
\newblock Point2Pix: Photo-Realistic Point Cloud Rendering via Neural Radiance
  Fields.
\newblock In \emph{Proceedings of the IEEE/CVF Conference on Computer Vision
  and Pattern Recognition}, 8349--8358.

\bibitem[{Jayasundara et~al.(2023)Jayasundara, Agrawal, Heron, Shrivastava, and
  Davis}]{jayasundara2023flexnerf}
Jayasundara, V.; Agrawal, A.; Heron, N.; Shrivastava, A.; and Davis, L.~S.
  2023.
\newblock FlexNeRF: Photorealistic free-viewpoint rendering of moving humans
  from sparse views.
\newblock In \emph{Proceedings of the IEEE/CVF Conference on Computer Vision
  and Pattern Recognition}, 21118--21127.

\bibitem[{Jiang et~al.(2022)Jiang, Yi, Samei, Tuzel, and
  Ranjan}]{jiang2022neuman}
Jiang, W.; Yi, K.~M.; Samei, G.; Tuzel, O.; and Ranjan, A. 2022.
\newblock Neuman: Neural human radiance field from a single video.
\newblock In \emph{European Conference on Computer Vision}, 402--418. Springer.

\bibitem[{Kwon et~al.(2021)Kwon, Kim, Ceylan, and Fuchs}]{kwon2021neural}
Kwon, Y.; Kim, D.; Ceylan, D.; and Fuchs, H. 2021.
\newblock Neural human performer: Learning generalizable radiance fields for
  human performance rendering.
\newblock \emph{Advances in Neural Information Processing Systems}, 34:
  24741--24752.

\bibitem[{Liu et~al.(2023{\natexlab{a}})Liu, Zhang, Zheng, and
  Duan}]{liu2023semantic}
Liu, F.; Zhang, C.; Zheng, Y.; and Duan, Y. 2023{\natexlab{a}}.
\newblock Semantic Ray: Learning a Generalizable Semantic Field with
  Cross-Reprojection Attention.
\newblock In \emph{Proceedings of the IEEE/CVF Conference on Computer Vision
  and Pattern Recognition}, 17386--17396.

\bibitem[{Liu et~al.(2021)Liu, Habermann, Rudnev, Sarkar, Gu, and
  Theobalt}]{liu2021neural}
Liu, L.; Habermann, M.; Rudnev, V.; Sarkar, K.; Gu, J.; and Theobalt, C. 2021.
\newblock Neural actor: Neural free-view synthesis of human actors with pose
  control.
\newblock \emph{ACM transactions on graphics (TOG)}, 40(6): 1--16.

\bibitem[{Liu et~al.(2020)Liu, Xu, Habermann, Zollh{\"o}fer, Bernard, Kim,
  Wang, and Theobalt}]{liu2020neural}
Liu, L.; Xu, W.; Habermann, M.; Zollh{\"o}fer, M.; Bernard, F.; Kim, H.; Wang,
  W.; and Theobalt, C. 2020.
\newblock Neural human video rendering by learning dynamic textures and
  rendering-to-video translation.
\newblock \emph{arXiv preprint arXiv:2001.04947}.

\bibitem[{Liu et~al.(2023{\natexlab{b}})Liu, Gao, Meuleman, Tseng, Saraf, Kim,
  Chuang, Kopf, and Huang}]{liu2023robust}
Liu, Y.-L.; Gao, C.; Meuleman, A.; Tseng, H.-Y.; Saraf, A.; Kim, C.; Chuang,
  Y.-Y.; Kopf, J.; and Huang, J.-B. 2023{\natexlab{b}}.
\newblock Robust Dynamic Radiance Fields.
\newblock \emph{arXiv preprint arXiv:2301.02239}.

\bibitem[{Loper et~al.(2015)Loper, Mahmood, Romero, Pons-Moll, and
  Black}]{loper2015smpl}
Loper, M.; Mahmood, N.; Romero, J.; Pons-Moll, G.; and Black, M.~J. 2015.
\newblock SMPL: A skinned multi-person linear model.
\newblock \emph{ACM transactions on graphics (TOG)}, 34(6): 1--16.

\bibitem[{Mildenhall et~al.(2020)Mildenhall, Srinivasan, Tancik, Barron,
  Ramamoorthi, and Ng}]{DBLP:journals/corr/abs-2003-08934}
Mildenhall, B.; Srinivasan, P.~P.; Tancik, M.; Barron, J.~T.; Ramamoorthi, R.;
  and Ng, R. 2020.
\newblock NeRF: Representing Scenes as Neural Radiance Fields for View
  Synthesis.
\newblock \emph{CoRR}, abs/2003.08934.

\bibitem[{Niemeyer and Geiger(2021)}]{niemeyer2021giraffe}
Niemeyer, M.; and Geiger, A. 2021.
\newblock Giraffe: Representing scenes as compositional generative neural
  feature fields.
\newblock In \emph{Proceedings of the IEEE/CVF Conference on Computer Vision
  and Pattern Recognition}, 11453--11464.

\bibitem[{Peng et~al.(2021{\natexlab{a}})Peng, Dong, Wang, Zhang, Shuai, Zhou,
  and Bao}]{peng2021animatable}
Peng, S.; Dong, J.; Wang, Q.; Zhang, S.; Shuai, Q.; Zhou, X.; and Bao, H.
  2021{\natexlab{a}}.
\newblock Animatable neural radiance fields for modeling dynamic human bodies.
\newblock In \emph{Proceedings of the IEEE/CVF International Conference on
  Computer Vision}, 14314--14323.

\bibitem[{Peng et~al.(2021{\natexlab{b}})Peng, Zhang, Xu, Wang, Shuai, Bao, and
  Zhou}]{peng2021neural}
Peng, S.; Zhang, Y.; Xu, Y.; Wang, Q.; Shuai, Q.; Bao, H.; and Zhou, X.
  2021{\natexlab{b}}.
\newblock Neural body: Implicit neural representations with structured latent
  codes for novel view synthesis of dynamic humans.
\newblock In \emph{Proceedings of the IEEE/CVF Conference on Computer Vision
  and Pattern Recognition}, 9054--9063.

\bibitem[{Qi et~al.(2017)Qi, Yi, Su, and Guibas}]{qi2017pointnet++}
Qi, C.~R.; Yi, L.; Su, H.; and Guibas, L.~J. 2017.
\newblock Pointnet++: Deep hierarchical feature learning on point sets in a
  metric space.
\newblock \emph{Advances in neural information processing systems}, 30.

\bibitem[{Shao et~al.(2022)Shao, Zhang, Zhang, Chen, Cao, Yu, and
  Liu}]{shao2022doublefield}
Shao, R.; Zhang, H.; Zhang, H.; Chen, M.; Cao, Y.-P.; Yu, T.; and Liu, Y. 2022.
\newblock Doublefield: Bridging the neural surface and radiance fields for
  high-fidelity human reconstruction and rendering.
\newblock In \emph{Proceedings of the IEEE/CVF Conference on Computer Vision
  and Pattern Recognition}, 15872--15882.

\bibitem[{Srinivasan et~al.(2021)Srinivasan, Deng, Zhang, Tancik, Mildenhall,
  and Barron}]{srinivasan2021nerv}
Srinivasan, P.~P.; Deng, B.; Zhang, X.; Tancik, M.; Mildenhall, B.; and Barron,
  J.~T. 2021.
\newblock Nerv: Neural reflectance and visibility fields for relighting and
  view synthesis.
\newblock In \emph{Proceedings of the IEEE/CVF Conference on Computer Vision
  and Pattern Recognition}, 7495--7504.

\bibitem[{Tseng et~al.(2022)Tseng, Liao, Yen-Chen, and Sun}]{tseng2022cla}
Tseng, W.-C.; Liao, H.-J.; Yen-Chen, L.; and Sun, M. 2022.
\newblock Cla-nerf: Category-level articulated neural radiance field.
\newblock In \emph{2022 International Conference on Robotics and Automation
  (ICRA)}, 8454--8460. IEEE.

\bibitem[{Vora et~al.(2021)Vora, Radwan, Greff, Meyer, Genova, Sajjadi, Pot,
  Tagliasacchi, and Duckworth}]{vora2021nesf}
Vora, S.; Radwan, N.; Greff, K.; Meyer, H.; Genova, K.; Sajjadi, M.~S.; Pot,
  E.; Tagliasacchi, A.; and Duckworth, D. 2021.
\newblock Nesf: Neural semantic fields for generalizable semantic segmentation
  of 3d scenes.
\newblock \emph{arXiv preprint arXiv:2111.13260}.

\bibitem[{Weng, Curless, and Kemelmacher-Shlizerman(2020)}]{weng2020vid2actor}
Weng, C.-Y.; Curless, B.; and Kemelmacher-Shlizerman, I. 2020.
\newblock Vid2actor: Free-viewpoint animatable person synthesis from video in
  the wild.
\newblock \emph{arXiv preprint arXiv:2012.12884}.

\bibitem[{Weng et~al.(2022)Weng, Curless, Srinivasan, Barron, and
  Kemelmacher-Shlizerman}]{weng2022humannerf}
Weng, C.-Y.; Curless, B.; Srinivasan, P.~P.; Barron, J.~T.; and
  Kemelmacher-Shlizerman, I. 2022.
\newblock Humannerf: Free-viewpoint rendering of moving people from monocular
  video.
\newblock In \emph{Proceedings of the IEEE/CVF Conference on Computer Vision
  and Pattern Recognition}, 16210--16220.

\bibitem[{Weng et~al.(2023)Weng, Srinivasan, Curless, and
  Kemelmacher-Shlizerman}]{weng2023personnerf}
Weng, C.-Y.; Srinivasan, P.~P.; Curless, B.; and Kemelmacher-Shlizerman, I.
  2023.
\newblock PersonNeRF: Personalized Reconstruction from Photo Collections.
\newblock In \emph{Proceedings of the IEEE/CVF Conference on Computer Vision
  and Pattern Recognition}, 524--533.

\bibitem[{Xu et~al.(2022)Xu, Xu, Philip, Bi, Shu, Sunkavalli, and
  Neumann}]{xu2022point}
Xu, Q.; Xu, Z.; Philip, J.; Bi, S.; Shu, Z.; Sunkavalli, K.; and Neumann, U.
  2022.
\newblock Point-nerf: Point-based neural radiance fields.
\newblock In \emph{Proceedings of the IEEE/CVF Conference on Computer Vision
  and Pattern Recognition}, 5438--5448.

\bibitem[{Yu et~al.(2023)Yu, Cheng, Liu, Wu, and Lin}]{yu2023monohuman}
Yu, Z.; Cheng, W.; Liu, X.; Wu, W.; and Lin, K.-Y. 2023.
\newblock MonoHuman: Animatable Human Neural Field from Monocular Video.
\newblock In \emph{Proceedings of the IEEE/CVF Conference on Computer Vision
  and Pattern Recognition}, 16943--16953.

\bibitem[{Zhang et~al.(2020)Zhang, Riegler, Snavely, and
  Koltun}]{zhang2020nerf++}
Zhang, K.; Riegler, G.; Snavely, N.; and Koltun, V. 2020.
\newblock Nerf++: Analyzing and improving neural radiance fields.
\newblock \emph{arXiv preprint arXiv:2010.07492}.

\bibitem[{Zhang et~al.(2021)Zhang, Srinivasan, Deng, Debevec, Freeman, and
  Barron}]{zhang2021nerfactor}
Zhang, X.; Srinivasan, P.~P.; Deng, B.; Debevec, P.; Freeman, W.~T.; and
  Barron, J.~T. 2021.
\newblock Nerfactor: Neural factorization of shape and reflectance under an
  unknown illumination.
\newblock \emph{ACM Transactions on Graphics (ToG)}, 40(6): 1--18.

\bibitem[{Zhao et~al.(2022)Zhao, Yang, Zhang, Lin, Zhang, Yu, and
  Xu}]{zhao2022humannerf}
Zhao, F.; Yang, W.; Zhang, J.; Lin, P.; Zhang, Y.; Yu, J.; and Xu, L. 2022.
\newblock Humannerf: Efficiently generated human radiance field from sparse
  inputs.
\newblock In \emph{Proceedings of the IEEE/CVF Conference on Computer Vision
  and Pattern Recognition}, 7743--7753.

\bibitem[{Zheng et~al.(2022)Zheng, Huang, Yu, Zhang, Guo, and
  Liu}]{zheng2022structured}
Zheng, Z.; Huang, H.; Yu, T.; Zhang, H.; Guo, Y.; and Liu, Y. 2022.
\newblock Structured local radiance fields for human avatar modeling.
\newblock In \emph{Proceedings of the IEEE/CVF Conference on Computer Vision
  and Pattern Recognition}, 15893--15903.

\bibitem[{Zhi et~al.(2021)Zhi, Laidlow, Leutenegger, and
  Davison}]{zhi2021place}
Zhi, S.; Laidlow, T.; Leutenegger, S.; and Davison, A.~J. 2021.
\newblock In-place scene labelling and understanding with implicit scene
  representation.
\newblock In \emph{Proceedings of the IEEE/CVF International Conference on
  Computer Vision}, 15838--15847.

\end{thebibliography}
\end{document}


\maketitle

\begin{abstract}
The neural rendering of humans is a topic of great research value. However, previous works mostly focus on achieving photorealistic detail, neglecting the exploration of human parsing with semantic labels, which is important to human-centric analysis and image editing. Human Parsing is intrinsically related to radiance reconstruction, as similar appearance and geometry often correspond to similar semantic parts. Previous researches often design a motion field that maps from the observation space to the canonical space, while it exhibits tendencies of either underfitting or overfitting, resulting in limited generalization. In this paper, we present a novel method, termed Semantic-Human, that achieves both photorealistic details and viewpoint-consistent human parsing for the neural rendering of humans. Specifically, we extend neural radiance fields (NeRF) to jointly encode semantics, appearance and geometry to achieve accurate 2D semantic labels using noisy pseudo-label supervision. By means of the inherent consistency and smoothness properties of NeRF, Semantic-Human effectively achieves consistent human parsing in both continuous and novel views. We also introduce constraints to the motion field from the SMPL surface and regularization to the recovered volumetric geometry. We have evaluated it on the ZJU-MoCap dataset, and highly competitive results have demonstrated the effectiveness of our proposed Semantic-Human. We also showcase various compelling applications, including label denoising, label synthesis and image editing, and empirically validate its advantageous properties.
\end{abstract}

\section{Introduction}

\bibliography{aaai23}